%% file: acl_latex.tex
\documentclass[11pt]{article}

\usepackage[final]{acl}
\usepackage{amsmath} 
\usepackage{enumitem}
\usepackage{times}
\usepackage{caption}
\usepackage{multirow} 
\usepackage{latexsym}
\usepackage{booktabs} 
\usepackage[T1]{fontenc}

\usepackage{CJKutf8}

\usepackage{color}
\usepackage{tcolorbox}          
\tcbuselibrary{skins,breakable} 
\usepackage[table]{xcolor}

\definecolor{tred}{RGB}{251, 130, 132}
\newcommand{\highlightbox}[3]{
\begin{tcolorbox}[
    enhanced jigsaw,
    breakable,
    pad at break*=1mm,
    colback=#1!5!white,
    colframe=#1,
    title=#2
]
#3
\end{tcolorbox}
}

\usepackage[utf8]{inputenc}

\usepackage{microtype}

\usepackage{inconsolata}

\usepackage{graphicx}

\title{Reinforcement Learning with Semantic Rewards Enables Low-Resource Language Expansion without Alignment Tax}

\author{%
  \small
  \begin{tabular}[t]{@{}c@{}}
    Zeli Su\textsuperscript{1,2$\ast$} \quad
    Ziyin Zhang\textsuperscript{3$\ast$} \quad
    Zhou Liu\textsuperscript{5$\ast$} \quad
    Xuexian Song\textsuperscript{6} \quad
    Zhankai Xu\textsuperscript{2} \\
    Longfei Zheng\textsuperscript{2} \quad
    Xiaolu Zhang\textsuperscript{2} \quad
    Rong Fu\textsuperscript{4} \quad
    Guixian Xu\textsuperscript{1,7$\dagger$} \quad
    Wentao Zhang\textsuperscript{5$\dagger$}
  \end{tabular}
  \\
  \vspace{4pt}
  \footnotesize
  \begin{tabular}[t]{@{}c@{}}
    \textsuperscript{1} Minzu University of China \quad
    \textsuperscript{2} Ant Group \quad
    \textsuperscript{3} Shanghai Jiao Tong University \quad
    \textsuperscript{4} University of Macau \quad
    \textsuperscript{5} Peking University \\
    \textsuperscript{6} Institute of Automation, Chinese Academy of Sciences \quad
    \textsuperscript{7} Hainan International College, Minzu University of China \\
  \end{tabular}
  \\
  \vspace{2pt}
  \footnotesize\texttt{\{rickamorty, guixian\_xu\}@muc.edu.cn} \quad
  \footnotesize\texttt{daenerystargaryen@sjtu.edu.cn} \\
  \footnotesize\texttt{zhouliu25@stu.pku.edu.cn} \quad
    \footnotesize\texttt{songxuexian5@gmail.com} \quad
  \footnotesize\texttt{\{xuzhankai.xzk, zlf206411\}@antgroup.com} \quad \\
  \footnotesize\texttt{yueyin.zxl@antfin.com} \quad
  \footnotesize\texttt{mc46603@um.edu.mo}  \quad
  \footnotesize\texttt{wentao.zhang@pku.edu.cn} \quad 
}

\begin{document}
\maketitle

\begin{abstract}
Extending large language models (LLMs) to low-resource languages often incurs an ``alignment tax'': improvements in the target language come at the cost of catastrophic forgetting in general capabilities. We argue that this trade-off arises from the rigidity of supervised fine-tuning (SFT), which enforces token-level surface imitation on narrow and biased data distributions. To address this limitation, we propose a semantic-space alignment paradigm powered by \textbf{Group Relative Policy Optimization (GRPO)}, where the model is optimized using \textbf{embedding-level semantic rewards} rather than likelihood maximization. This objective encourages meaning preservation through flexible realizations, enabling controlled updates that reduce destructive interference with pretrained knowledge. We evaluate our approach on Tibetan--Chinese machine translation and Tibetan headline generation. Experiments show that our method acquires low-resource capabilities while markedly mitigating alignment tax, preserving general competence more effectively than SFT. Despite producing less rigid surface overlap, semantic RL yields higher semantic quality and preference in open-ended generation, and few-shot transfer results indicate that it learns more transferable and robust representations under limited supervision. Overall, our study demonstrates that reinforcement learning with semantic rewards provides a safer and more reliable pathway for inclusive low-resource language expansion.
\end{abstract}

\input{section/introduction}

\input{section/related_work}
\input{section/methods}

\input{section/Experiment}

\section{Conclusion}
This paper argues that low-resource language expansion should be treated as an \emph{alignment} problem, where the core objective is semantic consistency rather than token-level imitation. We propose a semantic-space alignment paradigm instantiated with reinforcement learning driven by embedding-level semantic similarity and a strict language-consistency constraint. Experiments on Tibetan--Chinese machine translation and Tibetan headline generation show that semantic-reward-driven RL acquires low-resource language capabilities while substantially reducing alignment tax, preserving dominant-language competence with near-zero forgetting. We further observe a consistent mismatch between reference-based metrics and semantic quality: despite weaker n-gram overlap, RL is often preferred by LLM-based judges in open-ended generation and yields more transferable representations under few-shot transfer.Taken together, these findings suggest that semantic-space alignment offers a scalable path for extending LLMs to weakly supported languages under data scarcity, shifting low-resource adaptation from distribution matching toward meaning-centered alignment.

\section*{Limitations}

While modern LLMs nominally support many languages, identifying a language with weak model performance often implies severe data scarcity, as in the case of Tibetan. The limited and domain-narrow nature of available data (e.g., translation corpora) may cause supervised fine-tuning to achieve artificially high in-domain metrics that do not fully reflect real-world generalization, making some degree of overfitting unavoidable.

\section*{Ethical Considerations}

This work promotes inclusive language modeling by extending LLMs to low-resource languages such as Tibetan. All data are publicly available or internally licensed and contain no personal or sensitive information. No human participants were involved, and all evaluations were performed automatically using LLM-based judges. While our method reduces overfitting and potential bias from narrow supervision, residual pretrained biases may persist. Future research should further assess fairness and bias in low-resource settings.

\section*{Acknowledgments}
This work was supported by the Hainan Provincial Joint Project of the Li'an International Education Innovation Pilot Zone (Grant No.~624LALH006).



\bibliography{custom.bib}

\input{appendix}

\end{document}

%% file: section/introduction.tex
\begin{figure}
    \centering
    \includegraphics[width=1\linewidth]{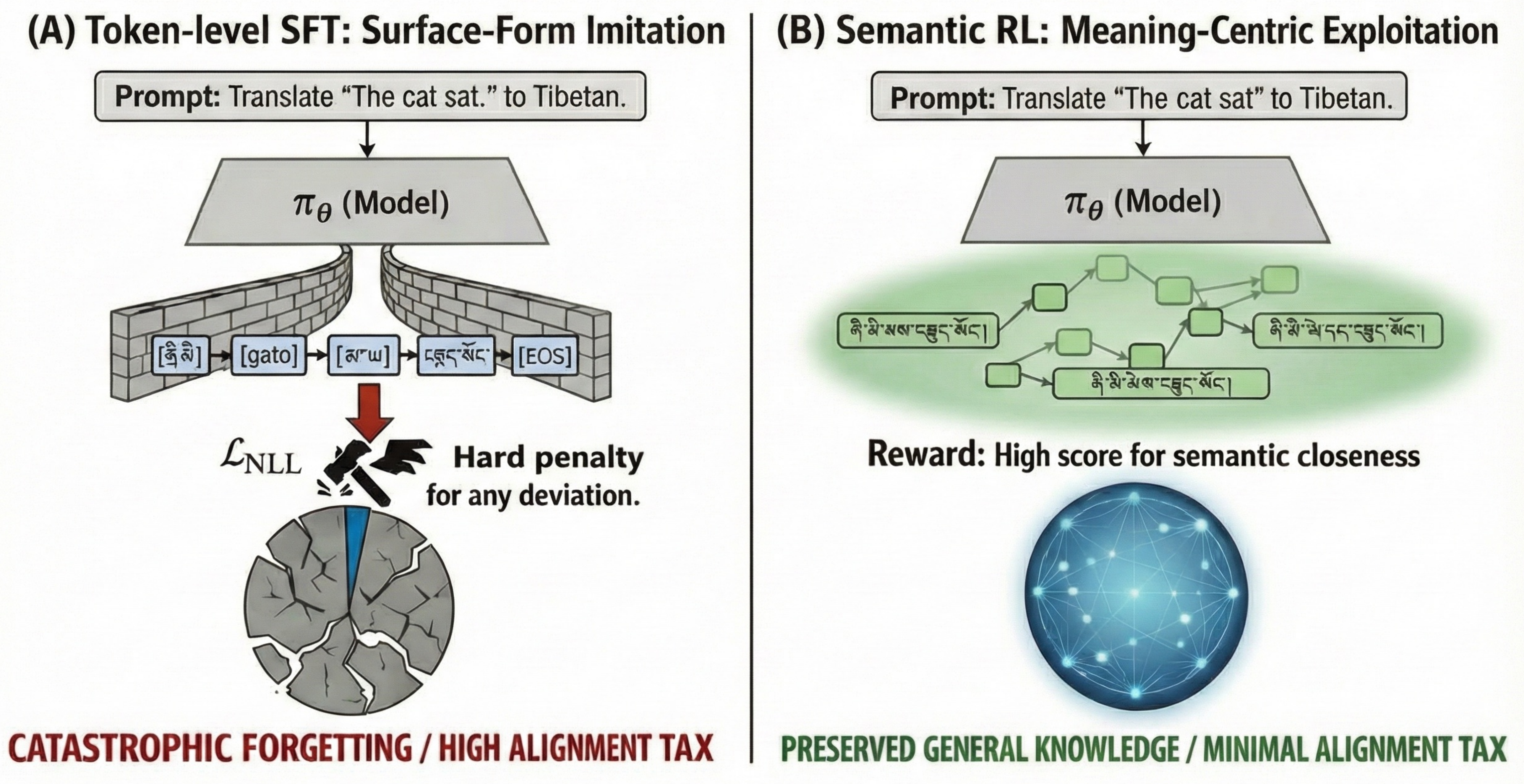}
    \caption{Token-level alignment versus semantic-space alignment in low-resource language expansion. 
Token-level supervised fine-tuning enforces surface-form imitation under teacher forcing, often causing catastrophic forgetting and high alignment tax. 
In contrast, semantic-space alignment optimizes meaning preservation with constrained reinforcement learning policy updates, allowing flexible realizations while preserving pretrained knowledge.}
    \label{fig:1_plot}

\end{figure}


\section{Introduction}

Large language models (LLMs) have achieved remarkable performance across a wide range of tasks and languages through large-scale pretraining and post-training alignment~\cite{deepseek2025v3_2,yang2025qwen3,kimiteam2025k2,openai2025gpt41,comanici2025gemini25}. However, their capabilities remain highly uneven across languages: many languages are weakly supported, particularly for generation and reasoning-intensive tasks that require semantic abstraction rather than surface pattern matching. Improving model performance in such low-resource language settings therefore remains an important and challenging problem.

A common approach to improving low-resource language performance is further training on language-specific data, including continual pretraining and supervised instruction fine-tuning. Despite differences in data sources and training protocols, these methods share a common optimization paradigm: they rely on teacher-forced learning with token-level likelihood objectives, aligning the model to a target data distribution through surface-form imitation. Figure~\ref{fig:1_plot} illustrates the contrast between token-level alignment via surface-form imitation and semantic-space alignment based on meaning preservation, highlighting how different objectives lead to fundamentally different update behaviors under data scarcity. Much recent work on language expansion and adaptation follows this paradigm, using supervised or weakly supervised finetuning to inject new linguistic capabilities into pretrained models \citep{language—adaption1,language—adaption3}. Related efforts based on continued pretraining or language-specific model construction similarly perform strong distribution-level updates on limited and domain-constrained data \citep{language—adaption2,language—adaption4}.

While effective under abundant and diverse supervision, token-level distribution matching becomes problematic in low-resource language settings. Available training data are often limited in size, narrow in domain, and distributionally biased. Optimizing likelihood on such data encourages overly confident and rigid parameter updates, amplifying overfitting and interfering with representations learned during pretraining. Empirically, this interference often manifests as catastrophic forgetting: improvements in the target language are accompanied by degradation in existing high-resource language capabilities, a phenomenon that has been systematically observed in multilingual fine-tuning and low-resource representation learning settings \citep{liu-niehues-2025-conditions,schmidt2025robust}.

We argue that this phenomenon is not merely an optimization artifact but a structural outcome of the alignment objective itself. When alignment is equated with surface-form imitation on narrow distributions, representational capacity is aggressively reallocated, leading to an \emph{alignment tax}: gains in low-resource language performance achieved at the expense of general competence. This issue persists regardless of the adaptation method (e.g., continual pretraining or instruction tuning) as long as the objective enforces token-level matching on sparse data \citep{yamaguchi2025mitigating}. 

Consequently, we propose shifting perspective to view language expansion not merely as adaptation, but as an \emph{alignment problem under sparse supervision}. To address this, we introduce a semantic-space alignment paradigm that prioritizes meaning preservation over rigid surface-form imitation. We operationalize this framework using \textbf{Group Relative Policy Optimization (GRPO)}~\citep{shao2024deepseekmath}, employing \textbf{embedding-level semantic similarity} as the primary reward signal. Unlike teacher-forced training, this approach encourages the model to explore diverse linguistic realizations that maintain semantic equivalence. Crucially, by optimizing based on relative rewards within a sampled group, our method inherently incorporates the stability constraints of trust-region optimization~\citep{schulman2017ppo,rafailov2023dpo}, enabling the acquisition of low-resource capabilities while strictly limiting the destructive interference typical of unconstrained likelihood maximization.

We evaluate our approach on Tibetan--Chinese machine translation (MT) and Tibetan headline generation (HG). Empirical results demonstrate that semantic-reward-driven GRPO achieves a superior trade-off between adaptation and preservation. In the MT task, our method substantially reduces alignment tax, outperforming the strong SFT baseline by \textbf{+5.15} points on the dominant-language CMRC benchmark. Similarly, in the HG task, despite lower n-gram overlap, our model is preferred by LLM-based judges with a \textbf{+16.1\%} higher win rate compared to SFT. These findings suggest that semantic-space alignment offers a safer and more robust paradigm for improving low-resource language performance under data scarcity.

In summary, the main contributions of this paper are:
\begin{itemize}[leftmargin=*, topsep=1pt, itemsep=1pt, parsep=0.5pt, partopsep=0.5pt]
    \item We propose a \textbf{semantic-space alignment paradigm} that utilizes Group Relative Policy Optimization (GRPO) with embedding-level rewards to decouple meaning preservation from surface-form imitation.
    \item We demonstrate that this approach virtually eliminates the \textbf{alignment tax}, enabling significant low-resource gains while maintaining the model's general capabilities and pretrained knowledge.
    \item We show that our method produces \textbf{semantically superior outputs} that are preferred by LLM judges over SFT baselines, despite having lower n-gram overlap with rigid references.
    \item We validate that semantic RL yields \textbf{more transferable representations}, as evidenced by stronger few-shot generalization to downstream tasks compared to supervised methods.
\end{itemize}

%% file: section/related_work.tex
\section{Related Work}
\label{sec:related_work}

\subsection{Low-Resource Language Adaptation and Expansion}

Post-training language expansion andadaptation typically start from a pretrained foundation model and further train on language-specific data. Prior work studies supervised fine-tuning or instruction tuning for transferring language capabilities and scaling with data and model size \citep{language—adaption1,language—adaption3}, as well as continued pretraining and language-specific model construction that emphasize corpus selection and composition \citep{language—adaption2,language—adaption4}. Across these settings, selective adaptation on narrow distributions can induce catastrophic forgetting, degrading performance on non-target languages or tasks \citep{liu-niehues-2025-conditions,schmidt2025robust}.

Several mitigation strategies focus on constraining update magnitude or protecting previously learned capabilities, such as parameter-efficient finetuning (e.g., LoRA-style low-rank updates) and source-shielded adaptation \citep{lora_review,yamaguchi2025mitigating}. While these methods can reduce interference, they generally retain teacher-forced token-level likelihood objectives. Our work is complementary: we reconsider the alignment objective itself by optimizing semantic consistency via embedding-level rewards, aiming to improve weak-language capability with lower alignment tax.


\subsection{Reinforcement Learning for LLM Alignment}
Reinforcement learning (RL) is commonly used in LLM alignment when optimization objectives are sequence-level or non-differentiable, enabling learning beyond token-level supervised imitation \citep{christiano2017deep,ouyang2022training}. A core advantage of RL-based alignment methods is the use of constrained policy updates, such as trust-region or KL-regularized optimization, which limits drift from pretrained representations and improves stability \citep{schulman2015trpo,schulman2017ppo}. 

Recent variants retain this constrained-update principle while improving efficiency or flexibility, including direct preference optimization~\citep{rafailov2023dpo} and group-based policy optimization methods \citep{shao2024deepseekmath}, which we adopt in this work to enable controlled alignment towards semantic-level objectives.

%% file: section/methods.tex
\section{Method}

\subsection{Problem Formulation: Semantic-Space Alignment}

We study low-resource language expansion as an alignment problem. Given a pretrained instruction-following language model $\pi_{\text{base}}$, our goal is to acquire new capabilities in a low-resource language while preserving existing competencies in dominant, high-resource languages. 

Conventional supervised fine-tuning aligns models by maximizing token-level likelihood under a target data distribution. In low-resource settings, where the distribution is narrow and biased, this objective enforces surface-form imitation and often leads to overconfident updates and catastrophic forgetting. We instead frame alignment as \emph{semantic-space alignment}: model outputs are considered correct if they preserve meaning, regardless of their specific surface realization. This formulation explicitly decouples semantic adequacy from token-level matching and allows multiple valid expressions of the same intent.

Under this perspective, alignment is defined by semantic consistency rather than distribution matching. Our objective is therefore to optimize the model to produce outputs that are semantically equivalent to reference texts, while limiting interference with representations learned during pretraining.

\subsection{Two-Stage Training Paradigm}

To operationalize semantic-space alignment in low-resource settings, we adopt a two-stage training paradigm.

\paragraph{Stage 1: Cold-start supervised fine-tuning.}
We first perform a lightweight supervised fine-tuning step on a small subset of low-resource data to obtain an initial policy $\pi_{\text{init}}$. Specifically, we fine-tune the base model on 5k training instances for two epochs. The goal of this stage is not to achieve strong task performance, but to bootstrap minimal output competence in the target language, such as producing text in the correct script and maintaining basic language consistency. This cold-start initialization allows the model to reliably generate non-degenerate outputs in the low-resource language, ensuring that subsequent semantic rewards are meaningful and that reinforcement learning does not collapse into uninformative exploration.

\paragraph{Stage 2: Reinforcement learning with semantic rewards.}
Starting from $\pi_{\text{init}}$, we perform reinforcement learning to align the model in semantic space. In this stage, we utilize the remaining training data to drive learning through semantic rewards rather than token-level supervision. Reinforcement learning is conducted for a single epoch, during which the model is encouraged to explore diverse surface realizations while preserving semantic equivalence to reference texts. Constrained policy optimization is applied throughout training to control update magnitude, enabling the model to acquire low-resource language capabilities while minimizing destructive interference with pretrained representations.

\subsection{Reinforcement Learning for Semantic Alignment}

Optimizing semantic alignment is inherently a sequence-level problem over discrete outputs. The embedding-based semantic rewards used in our framework are not directly differentiable with respect to model parameters, making standard supervised learning objectives unsuitable. Reinforcement learning therefore provides a natural and principled framework for optimizing such non-differentiable, sequence-level objectives, allowing direct optimization of semantic consistency rather than token-level likelihood.

Beyond enabling optimization of semantic rewards, reinforcement learning also plays a critical role in preserving pretrained knowledge during low-resource adaptation. In contrast to supervised fine-tuning, which performs unconstrained likelihood maximization on narrow data distributions, constrained reinforcement learning methods explicitly limit policy updates. This controlled optimization is crucial for reducing destructive interference and mitigating catastrophic forgetting, making reinforcement learning particularly well-suited for semantic-space alignment under data scarcity.

We instantiate reinforcement learning using Group Relative Policy Optimization (GRPO, \citealp{shao2024deepseekmath}), a value-free variant of PPO-style constrained optimization. For each input prompt $x$, we sample a group of candidate outputs $\{y^{(k)}\}_{k=1}^{K}$ from the current policy $\pi_{\theta}(\cdot \mid x)$, compute their corresponding rewards, and update the policy based on relative comparisons within the group. GRPO inherits the key stabilization mechanisms of PPO, including trust-region-style constraints that limit policy drift between updates, while avoiding the need for an explicit value function. These properties make GRPO a practical and stable choice for semantic alignment, enabling effective learning from semantic rewards while maintaining existing language capabilities.

\subsection{Semantic Reward Design}

A central component of our framework is a semantic reward that explicitly defines the alignment objective for low-resource language expansion. Unlike supervised fine-tuning, which implicitly aligns models through token-level likelihood, our goal is to directly guide learning toward semantic consistency. The reward therefore serves not merely as an optimization signal, but as the primary mechanism that determines what the model is encouraged to learn.

\paragraph{Semantic embedding model and suitability for reinforcement learning.} \label{reward}
To instantiate the semantic reward, we employ a multilingual sentence-level embedding model trained under a contrastive bilingual alignment objective. The model is adapted on parallel sentence pairs, where translations are treated as positive examples and in-batch samples provide implicit negatives. Rather than training an encoder from scratch or enforcing surface-form similarity, this adaptation strategy emphasizes meaning preservation and fine-grained semantic discriminability across different linguistic realizations.

We empirically observe that bilingual contrastive training yields stronger semantic structure than monolingual adaptation, improving both cross-lingual alignment and intra-language separability. This property is particularly important for reinforcement learning, where the reward signal must reflect graded semantic differences rather than coarse topical similarity. As a sanity check, we construct a small diagnostic set of sentence pairs spanning different degrees of semantic equivalence and find that the embedding model assigns similarity scores that consistently correlate with these graded relationships. This indicates that the resulting embedding space is suitable for use as a semantic reward for guiding alignment in reinforcement learning.

\subsubsection{Embedding-Level Semantic Similarity Reward}

The primary learning signal in our framework is an embedding-level semantic similarity reward. Let $f(\cdot)$ denote the sentence embedding model described above, which maps text to normalized vector representations. Given a generated output $y$ and a reference text $y^{*}$, we compute their semantic similarity using cosine similarity:
\begin{equation}
s(y, y^{*}) = \cos \left( f(y), f(y^{*}) \right).
\end{equation}

This reward directly reflects our desired learning direction: outputs are encouraged to preserve meaning, regardless of surface realization. In contrast to token-level likelihood objectives, this formulation treats semantically equivalent paraphrases as equally valid and explicitly avoids overfitting to reference form. To stabilize optimization and focus learning on meaningful improvements beyond minimal adequacy, we apply a threshold-and-rescale shaping function:
\begin{equation}
R_{\text{sim}}(y, y^{*}) =
\begin{cases}
0, & s(y, y^{*}) \le \tau, \\
\frac{s(y, y^{*}) - \tau}{1 - \tau}, & s(y, y^{*}) > \tau,
\end{cases}
\end{equation}
where $\tau$ corresponds to a minimal semantic adequacy level achieved after cold-start fine-tuning. This shaping ensures that reinforcement learning primarily refines semantic quality rather than amplifying noise from low-quality generations.





\subsubsection{Language Consistency Reward}

Because the embedding model is multilingual, optimizing semantic similarity alone may
reward mixed-language or partially off-target outputs. To prevent this reward hacking
behavior, we introduce a language consistency reward based on a rule-based script check
using Unicode ranges and regular expressions:
\begin{equation}
R_{\text{lang}}(y) =
\begin{cases}
0, & \text{language mixed}, \\
1, & \text{language consistent}.
\end{cases}
\end{equation}
This acts as a hard constraint, ensuring that semantic optimization is carried out strictly
within the target low-resource language space.

\paragraph{Final reward.}
The final reward combines semantic similarity and language consistency:
\begin{equation}
R(y, y^{*}) = \lambda_{\text{sim}} R_{\text{sim}}(y, y^{*}) + \lambda_{\text{lang}} R_{\text{lang}}(y).
\end{equation}
Together, these components define our semantic alignment objective: the model is encouraged to improve semantic adequacy while being strictly constrained to produce linguistically consistent outputs in the low-resource language. In practice, we assign a larger weight to semantic similarity ($\lambda_{\text{sim}} = 1.5$) than to language consistency ($\lambda_{\text{lang}} = 1.0$), reflecting our design choice that semantic preservation constitutes the primary learning objective, while language consistency serves as a necessary constraint to prevent degenerate or off-language generations.

%% file: section/Experiment.tex
\section{Experiments}

We conduct a series of experiments to evaluate whether semantic-reward-driven reinforcement learning (RL) provides a better trade-off between low-resource language adaptation and preservation of existing capabilities compared to supervised fine-tuning (SFT). Our experiments are designed to answer three research questions: (1) whether RL effectively acquires low-resource language capabilities, (2) how RL and SFT differ in the trade-off between task performance and alignment tax, and (3) whether RL learns more transferable representations under data scarcity.

\subsection{Experimental Setup}

\label{sec:exp_setup}

\paragraph{Base model and adaptation.}
All experiments are conducted on \textbf{Qwen3-4B} with parameter-efficient fine-tuning via LoRA~\citep{lora}.
Unless otherwise specified, we apply LoRA to all linear projection layers in self-attention and MLP blocks.
We use a LoRA rank of $r=64$, scaling factor $\alpha=128$, and dropout rate of $0.05$.

\paragraph{Supervised fine-tuning (SFT).}
SFT is trained for three epochs in BF16 with a global batch size of 32,
using AdamW~\citep{adamw} with learning rate $2\times 10^{-5}$ and a cosine schedule (warmup ratio 0.1).

\paragraph{Semantic reward model.}
The semantic reward described in Section~\ref{reward} is instantiated using a bilingual sentence-embedding model built on top of CINO~\cite{cino}, a Tibetan-enhanced extension of XLM-R~\cite{xlm-r}. We adapt CINO into a sentence-level encoder using \textsc{SentenceTransformer}, and further specialize it on Chinese--Tibetan parallel data to produce embedding-based semantic similarity scores. The resulting encoder is used as a frozen reward model during RL and is not jointly optimized with the policy model.

\paragraph{Reinforcement learning (GRPO).}
Reinforcement learning is performed with GRPO~\cite{shao2024deepseekmath} starting from the SFT checkpoint,
trained for one epoch in BF16 with AdamW and learning rate $5\times 10^{-7}$ (effective global batch size 32).
For each prompt, we sample 8 candidates with temperature 0.8 and top-$p$ 0.9, using max prompt/completion lengths of 256 tokens.

\paragraph{Controlled comparison.}
This unified setup ensures that observed differences primarily reflect the alignment strategy (semantic-reward-driven RL vs.\ SFT),
rather than mismatched optimization or adaptation configurations.
\textbf{Training and hyperparameter details are provided in Appendix~\ref{app:training-details}.}

\subsubsection{Tasks and Datasets}

We evaluate our approach on two representative Tibetan low-resource generation tasks: cross-lingual machine translation (MT) and monolingual headline generation (HG).

\paragraph{Machine Translation (MT).}
For Tibetan--Chinese machine translation, we use an internal parallel corpus collected for training a vision--language model~\cite{wu2019large}. Specifically, the corpus consists of Tibetan--Chinese sentence pairs translated as part of the pretraining data construction pipeline for the VLM, rather than annotations produced by the model itself. We repurpose this parallel data as supervised training material for machine translation in our experiments. As the corpus was originally curated to support VLM pretraining, a large portion of the data is grounded in visual descriptions, resulting in a relatively narrow and domain-constrained distribution. While this dataset does not aim to be a comprehensive translation benchmark, it reflects a realistic low-resource scenario with limited domain diversity and is therefore suitable for studying alignment behavior under data scarcity.

\paragraph{Headline Generation (HG).}
For Tibetan headline generation, we use the Tibetan subset of the CMHG dataset~\cite{xu-etal-2025-cmhg}. Due to the fine-grained tokenization of Tibetan in the Qwen tokenizer, raw samples often result in excessively long sequences and high memory consumption. We therefore filter out samples exceeding 1024 tokens and retain shorter instances for both training and evaluation. After filtering, the dataset contains 16,449 training samples and 621 test samples, which are used consistently across all headline generation experiments.

\subsubsection{Evaluation Protocols}

We adopt a multi-dimensional evaluation protocol to capture both surface-level accuracy and semantic quality. For task performance, we report standard reference-based metrics (BLEU for MT and ROUGE for HG) as well as embedding-based semantic similarity. To assess semantic quality beyond reference matching, we conduct blind pairwise evaluations using an LLM-as-a-Judge, where judgments are produced by \textbf{GPT-5.2} under a fixed evaluation prompt; the full judging prompt and evaluation policy are provided in Appendix~\ref{app:judge-prompt}. To quantify alignment tax, we evaluate all models on a dominant-language benchmark (Chinese CMRC,~\citealp{cui-emnlp2019-cmrc2018}) before and after adaptation and report performance changes relative to the base model.

\subsection{Experiment 1: Effectiveness of Semantic-Reward RL} \label{Experiment-1}

We first evaluate whether semantic-reward-driven reinforcement learning (RL) effectively acquires low-resource language capabilities beyond minimal supervised initialization. Specifically, we compare RL against the cold-start SFT model on both Tibetan--Chinese machine translation (MT) and Tibetan headline generation (HG).

\paragraph{Machine Translation.}
Table~\ref{tab:exp1_mt} reports the results on Tibetan--Chinese MT. Starting from the same cold-start SFT checkpoint trained on 5k parallel sentence pairs, RL is further trained on approximately 90k additional samples using semantic rewards. Compared to the cold-start baseline, RL yields consistent improvements in both reference-based accuracy and semantic similarity. BLEU-4 increases from 0.3953 to 0.4519, while semantic similarity improves substantially from 0.5593 to 0.7164.

\begin{table}[t]
\centering
\begin{tabular}{lcc}
\hline
Model & BLEU-4 & Similarity \\
\hline
Cold-start SFT & 0.3953 & 0.5593 \\
RL (Ours) & \textbf{0.4519} & \textbf{0.7164} \\
\hline
\end{tabular}
\caption{Experiment 1 results on Tibetan--Chinese machine translation.}
\label{tab:exp1_mt}
\end{table}

\paragraph{Headline Generation.}
We observe a similar trend on Tibetan headline generation. As shown in Table~\ref{tab:exp1_hg}, the RL model trained on approximately 15k samples consistently outperforms the cold-start SFT baseline in both ROUGE-L and semantic similarity. In particular, ROUGE-L improves from 0.2204 to 0.2530, while semantic similarity increases from 0.5774 to 0.6404. 

\begin{table}[t]
\centering
\begin{tabular}{lcc}
\hline
Model & ROUGE-L & Similarity \\
\hline
Cold-start SFT & 0.2204 & 0.5774 \\
RL (Ours) & \textbf{0.2530} & \textbf{0.6404} \\
\hline
\end{tabular}
\caption{Experiment 1 results on Tibetan headline generation.}
\label{tab:exp1_hg}
\end{table}

\paragraph{Analysis.}
Across both translation and generation tasks, semantic-reward-driven RL consistently improves performance over the cold-start SFT baseline. Notably, the improvements are particularly pronounced in semantic similarity, suggesting that RL primarily refines meaning preservation rather than merely increasing surface-level overlap with references. These results confirm that embedding-level semantic rewards constitute a sufficiently informative alignment signal, enabling effective low-resource language learning beyond minimal supervised initialization.

\subsection{Experiment 2: Trade-off Between Task Performance and Alignment Tax}
In this experiment, we compare semantic-reward-driven RL with a Strong SFT baseline to characterize the trade-off between task performance and preservation of existing general capabilities (i.e., alignment tax). Unlike the cold-start SFT model used in Experiment~1 --- which is trained only on a small 5k subset of low-resource data and serves solely as the initialization policy for RL --- the \textbf{Strong SFT model is trained on the full available training data (i.e., the same combined dataset used by cold-start SFT + RL) under the same optimization and LoRA configuration}, representing the best-effort supervised adaptation outcome.
Table~\ref{tab:exp2_tradeoff} reports results on Tibetan--Chinese machine translation (MT) and Tibetan headline generation (HG), including task metrics, semantic similarity, and dominant-language performance on CMRC as a proxy for alignment tax. We additionally report LLM-based preference as a reference-free measure of semantic quality.

\begin{table*}[t]
\centering
\small
\setlength{\tabcolsep}{5pt}
\begin{tabular}{l|cc|cc|c}
\toprule
\textbf{Model} 
& \multicolumn{2}{c|}{\textbf{Task Performance}} 
& \multicolumn{2}{c|}{\textbf{General Capability (Alignment Tax)}} 
& \textbf{Semantic Quality} \\
& \textbf{Metric} & \textbf{Similarity} 
& \textbf{CMRC Avg} & \textbf{CMRC F1} 
& \textbf{LLM-Judge Win (\%)} \\
\midrule
\multicolumn{6}{c}{\textit{\textbf{Task 1: Tibetan--Chinese Machine Translation (MT)}}} \\
\midrule
Strong SFT 
& \textbf{0.6006} & \textbf{0.8282} 
& 41.82 & 62.99 
& 59.2 \\
RL (Ours) 
& 0.4519 & 0.7164 
& \textbf{46.97} & \textbf{65.79} 
& 33.5 \\
\cmidrule{1-6}
\textit{Gap (RL vs. SFT)} 
& \textcolor{red}{-0.1487} & \textcolor{red}{-0.1118} 
& \textcolor{blue}{\textbf{+5.15}} & \textcolor{blue}{\textbf{+2.80}} 
& \textcolor{red}{-25.7} \\
\midrule
\multicolumn{6}{c}{\textit{\textbf{Task 2: Tibetan Headline Generation (HG)}}} \\
\midrule
Strong SFT 
& \textbf{0.3095} & 0.6499 
& 44.20 & 65.30 
& 35.1 \\
RL (Ours) 
& 0.2530 & 0.6404 
& \textbf{45.10} & 65.20 
& \textbf{51.2} \\
\cmidrule{1-6}
\textit{Gap (RL vs. SFT)} 
& \textcolor{red}{-0.0565} & \textcolor{red}{-0.0095} 
& \textcolor{blue}{\textbf{+0.90}} & \textcolor{red}{-0.10} 
& \textcolor{blue}{\textbf{+16.1}} \\
\bottomrule
\end{tabular}
\caption{\textbf{Trade-off Analysis: Task Performance vs. Alignment Tax.} 
For machine translation, Strong SFT achieves higher task metrics but incurs a heavy alignment tax, reflected by a significant drop in CMRC performance. In contrast, RL preserves general language capabilities with substantially higher CMRC scores while sacrificing surface-level metrics. 
For headline generation, both methods exhibit comparable general capability preservation, but RL significantly outperforms SFT in semantic quality as measured by LLM-based judgment, despite lower n-gram-based scores.}
\label{tab:exp2_tradeoff}
\end{table*}

\begin{table}[t]
\centering
\begin{tabular}{lcc}
\hline
Initialization & ROUGE-L & Similarity \\
\hline
Base Model & 0.1585 & 0.4695 \\
MT-SFT & \textbf{0.1935} & 0.5456 \\
MT-RL (Ours) & 0.1918 & \textbf{0.5690} \\
\hline
\end{tabular}
\caption{Few-shot transfer from MT to HG with 1,000 HG training samples.}
\label{tab:fewshot}
\end{table}

\paragraph{Task 1: Tibetan--Chinese Machine Translation (MT).}
On MT, Strong SFT achieves higher reference-based scores, improving BLEU from 0.4519 (RL) to 0.6006 and semantic similarity from 0.7164 to 0.8282, reflecting stronger surface alignment to references. However, this advantage is less pronounced under LLM-based judgment: Strong SFT is preferred in 59.2\% of cases, while the RL-aligned model still wins 33.5\%, indicating competitive semantic quality despite lower BLEU.

\textbf{These metric gains come with a substantial alignment tax.} After adaptation, SFT suffers marked degradation on CMRC (41.82 Avg / 62.99 F1), whereas RL preserves general capability significantly better (46.97 Avg / 65.79 F1). Overall, token-level imitation inflates reference-based MT metrics at the cost of forgetting, while constrained semantic alignment via RL yields safer updates with lower alignment tax.



\paragraph{Task 2: Tibetan Headline Generation (HG).}
On HG, Strong SFT again achieves higher reference-based scores (ROUGE-L 0.3095 vs.\ 0.2530 for RL), while the semantic similarity gap remains small (0.6499 vs.\ 0.6404). Both methods largely preserve dominant-language performance, with only minor differences in CMRC. However, under LLM-based judgment, RL is strongly preferred: it wins 51.2\% of pairwise comparisons versus 35.1\% for SFT (+16.1 points). This suggests that in open-ended generation, semantic-reward-driven RL learns generation behaviors that go beyond reference imitation, capturing alternative yet semantically appropriate realizations that are more human-preferred despite lower n-gram overlap.

\paragraph{Overall analysis.}
Across both tasks, Table~\ref{tab:exp2_tradeoff} reveals a consistent pattern: supervised fine-tuning excels at maximizing reference-based metrics, while semantic-reward-driven RL better preserves general capabilities and improves semantic quality under preference-based evaluation. In MT, this trade-off manifests primarily as alignment tax, where SFT’s metric gains coincide with substantial forgetting. In HG, where multiple valid realizations exist, RL is consistently preferred by LLM judges despite lower ROUGE, indicating that it learns generation patterns not anchored to a single reference form.

Together, these results suggest a fundamental mismatch between reference-based metrics and true semantic quality in low-resource settings. By aligning models in semantic space rather than enforcing surface imitation, constrained RL enables the acquisition of alternative, semantically valid generation paradigms that are poorly reflected by n-gram metrics but better capture human preferences.

\subsection{Experiment 3: Few-Shot Transferability}

Finally, we examine whether the stronger MT metrics observed for SFT in Experiment~2 translate into better \emph{cross-task} generalization. While SFT achieves higher reference-based scores on MT, Experiment~2 also reveals a clear mismatch between such metrics and semantic quality, as reflected by alignment tax and LLM-based judgment. This raises a natural question: \emph{do higher MT scores indicate genuinely stronger Tibetan representations, or do they primarily capture task- and reference-specific surface patterns that are unlikely to transfer?} 

To answer this, we design a few-shot transfer test from MT to HG.

Concretely, we take the best MT checkpoints produced by Strong SFT and by RL, and fine-tune each of them on the HG task using only 1,000 training samples under identical training settings. This setting stresses representation reuse: with limited HG supervision, a model that learns more general and semantically grounded Tibetan features during MT should adapt more effectively than a model whose gains are dominated by task-specific imitation.

Table~\ref{tab:fewshot} reports the results. Both MT-adapted models improve substantially over the base model, confirming that MT training provides useful Tibetan signal for downstream generation. However, despite MT-SFT's strong MT performance, it does not retain a corresponding advantage in transfer. The RL-initialized model achieves a higher semantic similarity score (0.5690 vs.\ 0.5456), while maintaining comparable ROUGE-L (0.1918 vs.\ 0.1935). This indicates that the MT-SFT model's improvements are, at least in part, tied to MT-specific surface alignment and do not generalize as strongly to a different open-ended generation task, whereas semantic-reward-driven RL yields representations that transfer better under limited supervision.We further provide a mechanistic analysis of forgetting in Appendix~\ref{app:ood_analysis}, where we examine OOD token-level negative log-likelihood and KL divergence to the base model on a fixed CMRC evaluation set. The results are consistent with our main findings and suggest that semantic RL yields more controlled distributional adaptation than SFT.


Overall, this experiment supports our central claim that semantic-space alignment provides a safer and more effective adaptation paradigm in low-resource settings: while SFT can produce larger in-task metric gains, RL achieves more robust generalization across tasks, consistent with the practical needs of low-resource language expansion. To complement these downstream results, we further provide a mechanistic analysis of forgetting in Appendix~\ref{app:ood_analysis}, where we examine OOD token-level negative log-likelihood and KL divergence to the base model on a fixed CMRC evaluation set. The results are consistent with our main findings and suggest that semantic RL yields more controlled distributional adaptation than SFT.

\subsection{Reward Ablation}

To further understand the role of reward design under the same reinforcement learning framework, we conduct a reward ablation study on the unified Tibetan--Chinese machine translation task.
Specifically, under identical model, data, and training settings, we compare several reward combinations to isolate how different reward components affect semantic alignment performance.

\begin{table}[t]
\centering
\begin{tabular}{lc}
\toprule
Reward & Similarity \\
\midrule
Embedding + LC (Ours) & 0.7164 \\
BLEU + LC & 0.6375 \\
BLEU + Embedding + LC & 0.6175 \\
BLEU + Embedding & 0.2312 \\
\bottomrule
\end{tabular}
\caption{Reward ablation results on Tibetan--Chinese machine translation under matched settings. LC denotes the language consistency reward.}
\label{tab:reward_ablation}
\end{table}

Table~\ref{tab:reward_ablation} shows that the proposed reward combination, consisting of embedding similarity and language consistency, achieves the best semantic similarity among all configurations.
This result suggests that the performance gain does not come from reinforcement learning alone, but depends critically on how the reward is defined.

First, language consistency is necessary for stable semantic optimization in the multilingual setting.
Without language consistency, the model frequently produces mixed Tibetan--Chinese outputs during exploration, indicating that semantic similarity alone is insufficient to constrain generation into the target low-resource language space.

Second, BLEU-based rewards consistently weaken performance.
As a surface-form overlap objective, BLEU introduces token-level pressure that restricts semantic exploration and partially restores the rigidity of supervised imitation.
Even when combined with embedding reward and language consistency, it still degrades performance relative to the simpler Embedding + LC formulation.
This suggests that token-overlap rewards are not well aligned with the objective of semantic-space alignment, where multiple surface realizations may preserve the same meaning.

We also tested an additional length-constraint reward for translation.
However, it did not improve generation quality and reduced CMRC performance by approximately 2 points, indicating that excessive output constraints may further harm both capability retention and semantic alignment.
Overall, these results show that a lightweight semantic reward, together with a necessary target-language constraint, is more effective than stacking additional surface-form objectives.

%% file: appendix.tex
\clearpage
\appendix
\section{Training and Hyperparameter Details}
\label{app:training-details}

\subsection{Base model and LoRA configuration}
We adopt \textbf{Qwen3-4B-Instruct} as the base model and apply LoRA for all SFT and RL experiments.
Unless otherwise specified, LoRA adapters are inserted into:
\begin{itemize}
    \item \textbf{Self-attention projections:} \texttt{q\_proj}, \texttt{k\_proj}, \texttt{v\_proj}, \texttt{o\_proj}
    \item \textbf{MLP projections:} \texttt{gate\_proj}, \texttt{up\_proj}, \texttt{down\_proj}
\end{itemize}
We use LoRA rank $r=64$, scaling factor $\alpha=128$, and dropout 0.05 throughout all experiments.

\subsection{Supervised fine-tuning (SFT) details}
\begin{itemize}
    \item \textbf{Initialization:} Qwen3-4B-Instruct
    \item \textbf{Precision:} BF16
    \item \textbf{Training length:} 3 epochs
    \item \textbf{Optimizer:} AdamW
    \item \textbf{Learning rate:} $2 \times 10^{-5}$
    \item \textbf{Scheduler:} cosine decay with warmup ratio 0.1
    \item \textbf{Global batch size:} 32 (2 GPUs $\times$ per-device batch size 8 $\times$ gradient accumulation 2)
    \item \textbf{Sequence length:} determined by the data and model defaults, typically 1024--2048 tokens
\end{itemize}

\subsection{Reinforcement learning (GRPO) details}
We perform reinforcement learning using GRPO~\cite{shao2024deepseekmath} starting from the cold-start SFT checkpoint.
\begin{itemize}
    \item \textbf{Initialization:} SFT checkpoint (cold-start)
    \item \textbf{Precision:} BF16
    \item \textbf{Training length:} 1 epoch
    \item \textbf{Optimizer:} AdamW
    \item \textbf{Learning rate:} $5 \times 10^{-7}$
    \item \textbf{Effective global batch size:} 32 (per-device batch size 16, gradient accumulation 2)
    \item \textbf{Group size:} 8 sampled candidate generations per input prompt
    \item \textbf{Max prompt length:} 256 tokens
    \item \textbf{Max completion length:} 256 tokens
    \item \textbf{Sampling temperature:} 0.8
    \item \textbf{Nucleus sampling:} top-$p$ = 0.9
\end{itemize}

\subsection{Rationale for unified configuration}
Across SFT and RL, we keep the base model, LoRA configuration, precision, and batch scale as aligned as possible.
This reduces confounding factors and supports attributing performance differences to the alignment paradigm itself.

\section{LLM-judge Prompt}\label{app:judge-prompt}
\subsection{LLM-judge Prompt for Headline Generation Task}

For the Tibetan headline generation task (HG), we used prompt in Figure~\ref{fig:prompt-hg} to evaluate two candidate headlines generated for a Tibetan news article. The evaluation was conducted using GPT-5.2 as the LLM-judge.

\begin{figure*}[ht]
\highlightbox{tred}{LLM Judge Prompt for headline generation}{
You are an expert linguist specializing in Tibetan-Chinese journalism.

Your task is to evaluate two candidate headlines (Candidate 1 and Candidate 2) generated for a Tibetan news article.\\

\#\#\# Article:

[Source Text]:

\{src\}\\

[Candidate 1]:

\{cand\_1\}\\

[Candidate 2]:

\{cand\_2\}\\

\#\#\# Task:

Compare the two candidates.

- If Candidate 1 is significantly better, output: [[1]]

- If Candidate 2 is significantly better, output: [[2]]

- If both are equally good or bad, output: [[0]]\\

Provide a brief reason (in Chinese) before your decision.\\

\#\#\# Output Format:

Reason: <brief explanation>

Decision: [[1]] or [[2]] or [[0]]
}
\caption{Prompt for headline generation evaluation.}
\label{fig:prompt-hg}
\end{figure*}

\subsection{LLM-judge Prompt for Machine Translation Task}

For the Tibetan-Chinese machine translation task (MT), we used the prompt in Figure~\ref{fig:prompt-mt} to evaluate two candidate translations. This evaluation was also conducted using GPT-5.2 as the LLM-judge.

\begin{figure*}[ht]
\highlightbox{tred}{LLM Judge Prompt for machine translation}{
You are an expert linguist specializing in Tibetan-Chinese translation.

Your task is to evaluate two candidate translations (Candidate 1 and Candidate 2) for a Tibetan text.\\

\#\#\# Article:

[Source Text]:

\{src\}\\

[Candidate 1]:

\{cand\_1\}\\

[Candidate 2]:

\{cand\_2\}\\

\#\#\# Task:

Compare the two translations.

- If Candidate 1 is significantly better, output: [[1]]

- If Candidate 2 is significantly better, output: [[2]]

- If both are equally good or bad, output: [[0]]\\

Provide a brief reason (in English) before your decision.\\

\#\#\# Output Format:

Reason: <brief explanation>

Decision: [[1]] or [[2]] or [[0]]
}
\caption{Prompt for machine translation evaluation.}
\label{fig:prompt-mt}
\end{figure*}

In both tasks, the evaluation process was conducted blind, with no direct reference to the model outputs, ensuring an unbiased comparison of the candidate results.

\section{OOD Log-Likelihood and KL Divergence Analysis}
\label{app:ood_analysis}

To complement the downstream evaluation in the main text, we further analyze forgetting from a more mechanistic perspective using an out-of-distribution (OOD) evaluation set.
Specifically, we use passages from the Chinese Machine Reading Comprehension benchmark (CMRC) as a fixed OOD corpus without supervision signals, representing general language understanding ability outside the low-resource language-expansion training distribution.
On this set, we evaluate token-level negative log-likelihood (NLL) and KL divergence relative to the base model.
These measurements help characterize how different training strategies affect distributional drift beyond downstream task metrics.

\paragraph{OOD token-level negative log-likelihood and KL divergence.}
We first compare token-level NLL on the CMRC OOD set.
Lower NLL indicates that the adapted model remains closer to the base model's general language modeling behavior on unseen out-of-domain data.
We then examine KL divergence between the adapted models and the base model on the same OOD set, which provides a complementary view of distributional drift.

\begin{table*}[t]
\centering
\begin{tabular}{lcccc}
\toprule
\multicolumn{5}{c}{\textbf{(a) OOD token-level negative log-likelihood on the fixed CMRC evaluation set}} \\
\midrule
Model & Mean NLL & Median NLL & P10 NLL & P90 NLL \\
\midrule
Base Model & 2.6097 & 1.4285 & 0.3636 & 5.8844 \\
RL (final) & 2.8533 & 1.5543 & 0.3848 & 6.5000 \\
SFT (final) & 3.2504 & 1.7390 & 0.4636 & 7.3188 \\
\midrule
\multicolumn{5}{c}{} \\
\multicolumn{5}{c}{\textbf{(b) KL divergence to the base model on the fixed CMRC OOD set}} \\
\midrule
Model & Mean KL & Median KL & P10 KL & P90 KL \\
\midrule
SFT (cold-start) $\|$ Base & 0.0471 & 0.0099 & 0.0000 & 0.0912 \\
SFT (final) $\|$ Base & 0.0404 & 0.0113 & 0.0000 & 0.0932 \\
RL (final) $\|$ Base & 0.0410 & 0.0073 & 0.0000 & 0.0839 \\
\bottomrule
\end{tabular}
\caption{Mechanistic OOD analysis on a fixed CMRC evaluation set. Panel (a) reports token-level negative log-likelihood (NLL), where lower values indicate better preservation of the base model's out-of-domain language modeling behavior. Panel (b) reports KL divergence to the base model, characterizing distributional drift after adaptation.}
\label{tab:appendix_ood_analysis}
\end{table*}

As shown in Table~\ref{tab:appendix_ood_analysis}(a), RL leads to substantially smaller degradation on OOD data than SFT.
Compared to the base model, RL increases mean NLL by only +0.24, whereas SFT increases it by +0.64.
The difference is even more pronounced in the tail: the 90th-percentile NLL rises by +0.62 under RL but by +1.43 under SFT.
This suggests that forgetting under SFT disproportionately affects harder OOD examples, while semantic RL preserves more stable behavior across the distribution.

Table~\ref{tab:appendix_ood_analysis}(b) shows that RL and SFT have comparable mean KL divergence to the base model, indicating that the overall magnitude of adaptation is similar.
However, RL yields consistently lower median and tail KL values than SFT.
In particular, the 90th-percentile KL is lower for RL (0.0839) than for both cold-start SFT (0.0912) and final SFT (0.0932).
This suggests that semantic RL does not merely reduce the total amount of learning, but instead leads to more uniform distributional adaptation and avoids localized large shifts that are more characteristic of catastrophic forgetting.

\paragraph{Summary.}
Taken together, the OOD NLL and KL analyses provide complementary mechanistic evidence for the main findings in the paper.
Compared with supervised fine-tuning, semantic-reward RL induces smaller degradation in token-level likelihood on unseen OOD data, while also producing more controlled and less heavy-tailed divergence from the base model.
These observations support our interpretation that semantic RL mitigates alignment tax not by suppressing learning altogether, but by encouraging more uniform and less destructive adaptation.

%% file: custom.bib
@inproceedings{christiano2017deep,
  title        = {Deep Reinforcement Learning from Human Preferences},
  author       = {Christiano, Paul F. and Leike, Jan and Brown, Tom B. and Martic, Miljan and Legg, Shane and Amodei, Dario},
  booktitle    = {Advances in Neural Information Processing Systems},
  year         = {2017},
  url          = {https://arxiv.org/abs/1706.03741}
}

@inproceedings{ouyang2022training,
  title        = {Training Language Models to Follow Instructions with Human Feedback},
  author       = {Ouyang, Long and Wu, Jeff and Jiang, Xu and Almeida, Diogo and Wainwright, Carroll L. and Mishkin, Pamela and Zhang, Chong and Agarwal, Sandhini and Slama, Katarina and Ray, Alex and others},
  booktitle    = {Advances in Neural Information Processing Systems},
  year         = {2022},
  url          = {https://arxiv.org/abs/2203.02155}
}

@article{schulman2015trpo,
  title        = {Trust Region Policy Optimization},
  author       = {Schulman, John and Levine, Sergey and Moritz, Philipp and Jordan, Michael I. and Abbeel, Pieter},
  journal      = {arXiv preprint arXiv:1502.05477},
  year         = {2015},
  url          = {https://arxiv.org/abs/1502.05477}
}

@article{schulman2017ppo,
  title        = {Proximal Policy Optimization Algorithms},
  author       = {Schulman, John and Wolski, Filip and Dhariwal, Prafulla and Radford, Alec and Klimov, Oleg},
  journal      = {arXiv preprint arXiv:1707.06347},
  year         = {2017},
  url          = {https://arxiv.org/abs/1707.06347}
}

@article{rafailov2023dpo,
  title        = {Direct Preference Optimization: Your Language Model is Secretly a Reward Model},
  author       = {Rafailov, Rafael and Sharma, Archit and Mitchell, Eric and Ermon, Stefano and Manning, Christopher D. and Finn, Chelsea},
  journal      = {arXiv preprint arXiv:2305.18290},
  year         = {2023},
  url          = {https://arxiv.org/abs/2305.18290}
}

@article{shao2024deepseekmath,
  title        = {DeepSeekMath: Pushing the Limits of Mathematical Reasoning in Open Language Models},
  author       = {Shao, Zhihong and others},
  journal      = {arXiv preprint arXiv:2402.03300},
  year         = {2024},
  url          = {https://arxiv.org/abs/2402.03300}
}

@inproceedings{language—adaption1,
    title = "{S}amba{L}ingo: Teaching Large Language Models New Languages",
    author = "Csaki, Zoltan  and
      Li, Bo  and
      Li, Jonathan Lingjie  and
      Xu, Qiantong  and
      Pawakapan, Pian  and
      Zhang, Leon  and
      Du, Yun  and
      Zhao, Hengyu  and
      Hu, Changran  and
      Thakker, Urmish",
    editor = {S{\"a}lev{\"a}, Jonne  and
      Owodunni, Abraham},
    booktitle = "Proceedings of the Fourth Workshop on Multilingual Representation Learning (MRL 2024)",
    month = nov,
    year = "2024",
    address = "Miami, Florida, USA",
    publisher = "Association for Computational Linguistics",
    url = "https://aclanthology.org/2024.mrl-1.1/",
    doi = "10.18653/v1/2024.mrl-1.1",
    pages = "1--21"
}

@misc{language—adaption2,
      title={Curi\'o-Edu 7B: Examining Data Selection Impacts in LLM Continued Pretraining}, 
      author={Thales Sales Almeida and Rodrigo Nogueira and Hélio Pedrini},
      year={2025},
      eprint={2512.12770},
      archivePrefix={arXiv},
      primaryClass={cs.CL},
      url={https://arxiv.org/abs/2512.12770}, 
}

@article{language—adaption3,
  author       = {Jun Zhao and
                  Zhihao Zhang and
                  Luhui Gao and
                  Qi Zhang and
                  Tao Gui and
                  Xuanjing Huang},
  title        = {LLaMA Beyond English: An Empirical Study on Language Capability Transfer},
  journal      = {CoRR},
  volume       = {abs/2401.01055},
  year         = {2024},
  url          = {https://doi.org/10.48550/arXiv.2401.01055},
  doi          = {10.48550/ARXIV.2401.01055},
  eprinttype    = {arXiv},
  eprint       = {2401.01055},
  bibsource    = {dblp computer science bibliography, https://dblp.org}
}

@inproceedings{language—adaption4,
  title     = {ALLaM: Large Language Models for Arabic and English},
  author    = {Bari, M Saiful and Alnumay, Yazeed and Alzahrani, Norah A. and Alotaibi, Nouf M. and Alyahya, Hisham Abdullah and AlRashed, Sultan and Mirza, Faisal Abdulrahman and Alsubaie, Shaykhah Z. and Alahmed, Hassan A. and Alabduljabbar, Ghadah and Alkhathran, Raghad and Almushayqih, Yousef and Alnajim, Raneem and Alsubaihi, Salman and Al Mansour, Maryam and Hassan, Saad Amin and Alrubaian, Majed and Alammari, Ali and Alawami, Zaki and Al-Thubaity, Abdulmohsen and Abdelali, Ahmed and Kuriakose, Jeril and Abujabal, Abdalghani and Al-Twairesh, Nora and Alowisheq, Areeb and Khan, Haidar and others},
  booktitle = {International Conference on Learning Representations (ICLR) 2025},
  year      = {2025},
  note      = {Poster},
  url       = {https://openreview.net/forum?id=MscdsFVZrN}
}

@misc{deepseek2025v3_2,
  title         = {DeepSeek-V3.2: Pushing the Frontier of Open Large Language Models},
  author        = {DeepSeek-AI and others},
  year          = {2025},
  eprint        = {2512.02556},
  archivePrefix = {arXiv},
  primaryClass  = {cs.CL},
  doi           = {10.48550/arXiv.2512.02556},
  url           = {https://arxiv.org/abs/2512.02556}
}

@misc{yang2025qwen3,
  title         = {Qwen3 Technical Report},
  author        = {Yang, An and others},
  year          = {2025},
  eprint        = {2505.09388},
  archivePrefix = {arXiv},
  primaryClass  = {cs.CL},
  doi           = {10.48550/arXiv.2505.09388},
  url           = {https://arxiv.org/abs/2505.09388}
}

@misc{kimiteam2025k2,
  title         = {Kimi K2: Open Agentic Intelligence},
  author        = {Kimi Team and Bai, Yifan and others},
  year          = {2025},
  eprint        = {2507.20534},
  archivePrefix = {arXiv},
  primaryClass  = {cs.LG},
  doi           = {10.48550/arXiv.2507.20534},
  url           = {https://arxiv.org/abs/2507.20534}
}

@misc{openai2025gpt41,
  title        = {Introducing GPT-4.1 in the API},
  author       = {{OpenAI}},
  year         = {2025},
  month        = apr,
  howpublished = {OpenAI Blog},
  url          = {https://openai.com/index/gpt-4-1/},
  note         = {Accessed: 2025-12-25}
}

@misc{comanici2025gemini25,
  title         = {Gemini 2.5: Pushing the Frontier with Advanced Reasoning, Multimodality, Long Context, and Next Generation Agentic Capabilities},
  author        = {Comanici, Gheorghe and Bieber, Eric and Schaekermann, Mike and Pasupat, Ice and others},
  year          = {2025},
  eprint        = {2507.06261},
  archivePrefix = {arXiv},
  primaryClass  = {cs.CL},
  doi           = {10.48550/arXiv.2507.06261},
  url           = {https://arxiv.org/abs/2507.06261}
}

@inproceedings{liu-niehues-2025-conditions,
    title = {Conditions for Catastrophic Forgetting in Multilingual Translation},
    author = {Liu, Danni and Niehues, Jan},
    booktitle = {Proceedings of the 5th Workshop on Multilingual Representation Learning (MRL 2025)},
    month = nov,
    year = {2025},
    pages = {347--359},
    publisher = {Association for Computational Linguistics},
    doi = {10.18653/v1/2025.mrl-main.23},
    url = {https://aclanthology.org/2025.mrl-main.23/}
}

@misc{yamaguchi2025mitigating,
    title = {Mitigating Catastrophic Forgetting in Target Language Adaptation of LLMs via Source-Shielded Updates},
    author = {Yamaguchi, Atsuki and Morishita, Terufumi and Villavicencio, Aline and Aletras, Nikolaos},
    year = {2025},
    archivePrefix = {arXiv},
    eprint = {2512.04844},
    primaryClass = {cs.CL},
    url = {https://arxiv.org/abs/2512.04844}
}

@phdthesis{schmidt2025robust,

  title={Robust and Scalable Cross-Lingual Transfer},

  author={Schmidt, Fabian David},

  year={2025},

  school={Bayerische Julius-Maximilians-Universitaet Wuerzburg (Germany)}

}

@article{lora_review,
  author       = {Menglin Yang and
                  Jialin Chen and
                  Yifei Zhang and
                  Jiahong Liu and
                  Jiasheng Zhang and
                  Qiyao Ma and
                  Harshit Verma and
                  Qianru Zhang and
                  Min Zhou and
                  Irwin King and
                  Rex Ying},
  title        = {Low-Rank Adaptation for Foundation Models: {A} Comprehensive Review},
  journal      = {CoRR},
  volume       = {abs/2501.00365},
  year         = {2025},
  url          = {https://doi.org/10.48550/arXiv.2501.00365},
  doi          = {10.48550/ARXIV.2501.00365},
  eprinttype    = {arXiv},
  eprint       = {2501.00365},
  bibsource    = {dblp computer science bibliography, https://dblp.org}
}

@inproceedings{xu-etal-2025-cmhg,
    title = "{CMHG}: A Dataset and Benchmark for Headline Generation of Minority Languages in {C}hina",
    author = "Xu, Guixian  and
      Su, Zeli  and
      Zhang, Ziyin  and
      Liu, Jianing  and
      Han, Xu  and
      Zhang, Ting  and
      Dong, Yushuang",
    editor = "Christodoulopoulos, Christos  and
      Chakraborty, Tanmoy  and
      Rose, Carolyn  and
      Peng, Violet",
    booktitle = "Proceedings of the 2025 Conference on Empirical Methods in Natural Language Processing",
    month = nov,
    year = "2025",
    address = "Suzhou, China",
    publisher = "Association for Computational Linguistics",
    url = "https://aclanthology.org/2025.emnlp-main.622/",
    doi = "10.18653/v1/2025.emnlp-main.622",
    pages = "12350--12357",
    ISBN = "979-8-89176-332-6"
}

@inproceedings{wu2019large,
  title={Large-scale datasets for going deeper in image understanding},
  author={Wu, Jiahong and Zheng, He and Zhao, Bo and Li, Yixin and Yan, Baoming and Liang, Rui and Wang, Wenjia and Zhou, Shipei and Lin, Guosen and Fu, Yanwei and others},
  booktitle={2019 IEEE International Conference on Multimedia and Expo (ICME)},
  pages={1480--1485},
  year={2019},
  organization={IEEE}
}

@inproceedings{cui-emnlp2019-cmrc2018,
    title = "A Span-Extraction Dataset for {C}hinese Machine Reading Comprehension",
    author = "Cui, Yiming  and
      Liu, Ting  and
      Che, Wanxiang  and
      Xiao, Li  and
      Chen, Zhipeng  and
      Ma, Wentao  and
      Wang, Shijin  and
      Hu, Guoping",
    booktitle = "Proceedings of the 2019 Conference on Empirical Methods in Natural Language Processing and the 9th International Joint Conference on Natural Language Processing (EMNLP-IJCNLP)",
    month = nov,
    year = "2019",
    address = "Hong Kong, China",
    publisher = "Association for Computational Linguistics",
    url = "https://www.aclweb.org/anthology/D19-1600",
    doi = "10.18653/v1/D19-1600",
    pages = "5886--5891",
}

@inproceedings{cino,
    title = "{CINO}: A {C}hinese Minority Pre-trained Language Model",
    author = "Yang, Ziqing  and
      Xu, Zihang  and
      Cui, Yiming  and
      Wang, Baoxin  and
      Lin, Min  and
      Wu, Dayong  and
      Chen, Zhigang",
    booktitle = "Proceedings of the 29th International Conference on Computational Linguistics",
    month = oct,
    year = "2022",
    address = "Gyeongju, Republic of Korea",
    publisher = "International Committee on Computational Linguistics",
    url = "https://aclanthology.org/2022.coling-1.346",
    pages = "3937--3949"
}

@inproceedings{xlm-r,
    title = "Unsupervised Cross-lingual Representation Learning at Scale",
    author = "Conneau, Alexis  and
      Khandelwal, Kartikay  and
      Goyal, Naman  and
      Chaudhary, Vishrav  and
      Wenzek, Guillaume  and
      Guzm{\'a}n, Francisco  and
      Grave, Edouard  and
      Ott, Myle  and
      Zettlemoyer, Luke  and
      Stoyanov, Veselin",
    editor = "Jurafsky, Dan  and
      Chai, Joyce  and
      Schluter, Natalie  and
      Tetreault, Joel",
    booktitle = "Proceedings of the 58th Annual Meeting of the Association for Computational Linguistics",
    month = jul,
    year = "2020",
    address = "Online",
    publisher = "Association for Computational Linguistics",
    url = "https://aclanthology.org/2020.acl-main.747/",
    doi = "10.18653/v1/2020.acl-main.747",
    pages = "8440--8451"
}

@inproceedings{lora,
  author       = {Edward J. Hu and
                  Yelong Shen and
                  Phillip Wallis and
                  Zeyuan Allen{-}Zhu and
                  Yuanzhi Li and
                  Shean Wang and
                  Lu Wang and
                  Weizhu Chen},
  title        = {LoRA: Low-Rank Adaptation of Large Language Models},
  booktitle    = {The Tenth International Conference on Learning Representations, {ICLR}
                  2022, Virtual Event, April 25-29, 2022},
  publisher    = {OpenReview.net},
  year         = {2022},
  url          = {https://openreview.net/forum?id=nZeVKeeFYf9},
  bibsource    = {dblp computer science bibliography, https://dblp.org}
}

@inproceedings{adamw,
  author       = {Ilya Loshchilov and
                  Frank Hutter},
  title        = {Decoupled Weight Decay Regularization},
  booktitle    = {7th International Conference on Learning Representations, {ICLR} 2019,
                  New Orleans, LA, USA, May 6-9, 2019},
  publisher    = {OpenReview.net},
  year         = {2019},
  url          = {https://openreview.net/forum?id=Bkg6RiCqY7},
  bibsource    = {dblp computer science bibliography, https://dblp.org}
}
